\documentclass[letterpaper, 10 pt, journal, twoside]{IEEEtran}

\IEEEoverridecommandlockouts                              

\usepackage[dvipdfmx]{graphicx}

\usepackage{enumitem}
\usepackage{bm}
\usepackage{amsmath,amssymb}
\usepackage{mathtools}
\usepackage{physics}
\usepackage{listings}
\usepackage{url}
\usepackage{subfigmat}
\usepackage{color}
\usepackage{multirow} 
\usepackage[normalem]{ulem}
\usepackage{algorithm}
\usepackage{booktabs}
\usepackage{array}
\usepackage{makecell} 
\usepackage{algpseudocode}
\usepackage{xcolor}

\title{
SADP: Subgoal-Aware Diffusion Policy for Long-Horizon Manipulation Learned from Foundation Model Generated Demonstrations
}

\begin{document}

\markboth{IEEE Robotics and Automation Letters. VOL. 11, NO. 11, NOVEMBER 2026}
{Hu \MakeLowercase{\textit{et al.}}: SADP: Subgoal-Aware Diffusion Policy for Long-Horizon Manipulation Learned from Foundation Model Generated Demonstrations} 

\author{Site Hu$^{1}$ and Takato Horii$^{1}$
\thanks{Received 14 May 2026; accepted 10 September 2026. Date of publication
9 October 2026; date of current version 16 October 2026. This article was
recommended for publication by Associate Editor S. Zhang and Editor W. Pan
upon evaluation of the reviewers’ comments. This work was supported by the
JST Moonshot R\&D under Grant JPMJMS2011. (Corresponding author: Site
Hu.)}
\thanks{$^{1}$The authors are with the Department of Systems Innovation, Graduate School of Engineering Science, The University of Osaka, Osaka 565-0871, Japan
        {\tt\footnotesize (hu.site.es@osaka-u.ac.jp; takato@sys.es.osaka-u.ac.jp)}.}%
\thanks{This article has supplementary downloadable material available at
https://doi.org/10.1109/LRA.2026.3742285, provided by the authors.}
\thanks{Digital Object Identifier 10.1109/LRA.2026.3742285}
\thanks{%
\textcopyright{} 2026 IEEE. Personal use of this material
is permitted. Permission from IEEE must be obtained for
all other uses, in any current or future media, including
reprinting/republishing this material for advertising or
promotional purposes, creating new collective works,
for resale or redistribution to servers or lists, or reuse
of any copyrighted component of this work in other works.
}%
}

\maketitle

\begin{abstract}
Long-horizon robot manipulation requires policies to coordinate multiple intermediate subgoals and determine when to advance between them. However, many imitation learning methods are trained on task-level demonstrations without explicitly modeling the active subgoal, partly because standard robot learning datasets rarely provide subgoal-level supervision. Foundation-model-based demonstration generation already decomposes tasks into ordered natural-language subgoals, but this structure is typically not retained in the learned policy. To address this issue, this paper proposes Subgoal-Aware Diffusion Policy (SADP), a framework that retains the subgoal descriptions and trajectory boundaries produced during automatic demonstration generation and uses them as policy supervision without additional manual annotation. SADP conditions action generation on both task-level and subgoal-level descriptions, and a lightweight auxiliary head predicts a continuation score that triggers online subgoal switching. Experiments in RLBench simulations and real-world evaluations on a UR5e robot demonstrate that SADP achieves task success rates competitive with the strongest baseline. An offline evaluation on held-out generated demonstrations further shows that the continuation predictor recovers most subgoal boundaries defined by the generation pipeline, with errors dominated by delayed or missed transitions rather than premature ones. These results show that an explicit subgoal sequence and switching mechanism can be incorporated into a diffusion policy while retaining competitive task-level performance.
\end{abstract}

\begin{IEEEkeywords}
AI-Enabled Robotics; Learning from Demonstration; Deep Learning in Grasping and Manipulation
\end{IEEEkeywords}

\section{INTRODUCTION}

\IEEEPARstart{L}{ong-horizon} robot manipulation requires the coordinated execution of multiple intermediate subgoals, such as approaching, grasping, transporting, and placing objects~\cite{kroemer2021review}. However, many imitation learning (IL) policies are trained from task-level demonstrations and directly map observations and task instructions to low-level actions~\cite{correia2024survey}. The active subgoal and the decision of when to advance therefore remain implicit, limiting the use of stage-specific information for action generation and making execution progress difficult to monitor.

Existing approaches capture such a temporal structure in different ways. Hierarchical reinforcement learning (HRL) introduces latent subgoals or options to improve exploration and temporal abstraction in long-horizon control~\cite{hutsebaut2022hierarchical}. However, these representations serve mainly as internal abstractions for control rather than explicit natural-language conditions for learned visuomotor policies. Completion-aware methods such as SeqVLA~\cite{yang2025seqvla} explicitly condition action generation on subgoals and predict their completion for online switching, but require separately collected subgoal demonstrations and manually annotated frame-level continuation labels. SCIL~\cite{ouyang2025scil} similarly combines a stage observer trained on stage-labeled demonstrations with a stage-conditioned policy. Explicit subgoal-conditioned control and online switching therefore remain difficult to learn without costly subgoal-level supervision.

A major obstacle is that commonly used robot-learning datasets typically provide only task-level descriptions and sparse success labels, without explicit subgoal descriptions or transition supervision. Foundation models offer an alternative by decomposing task instructions into ordered subgoals and synthesizing demonstrations from high-level instructions~\cite{ha2023scaling, hu2025tarad}. Existing approaches, however, either use these subgoals only during planning and demonstration generation or distill the generated trajectories into task-level policies without retaining the active subgoal and transition structure. This leaves a gap between automatic subgoal-structured demonstration generation and learned policies that retain and exploit this structure during execution.

\begin{figure}[t]
    \centering
    \includegraphics[width=\columnwidth, keepaspectratio]{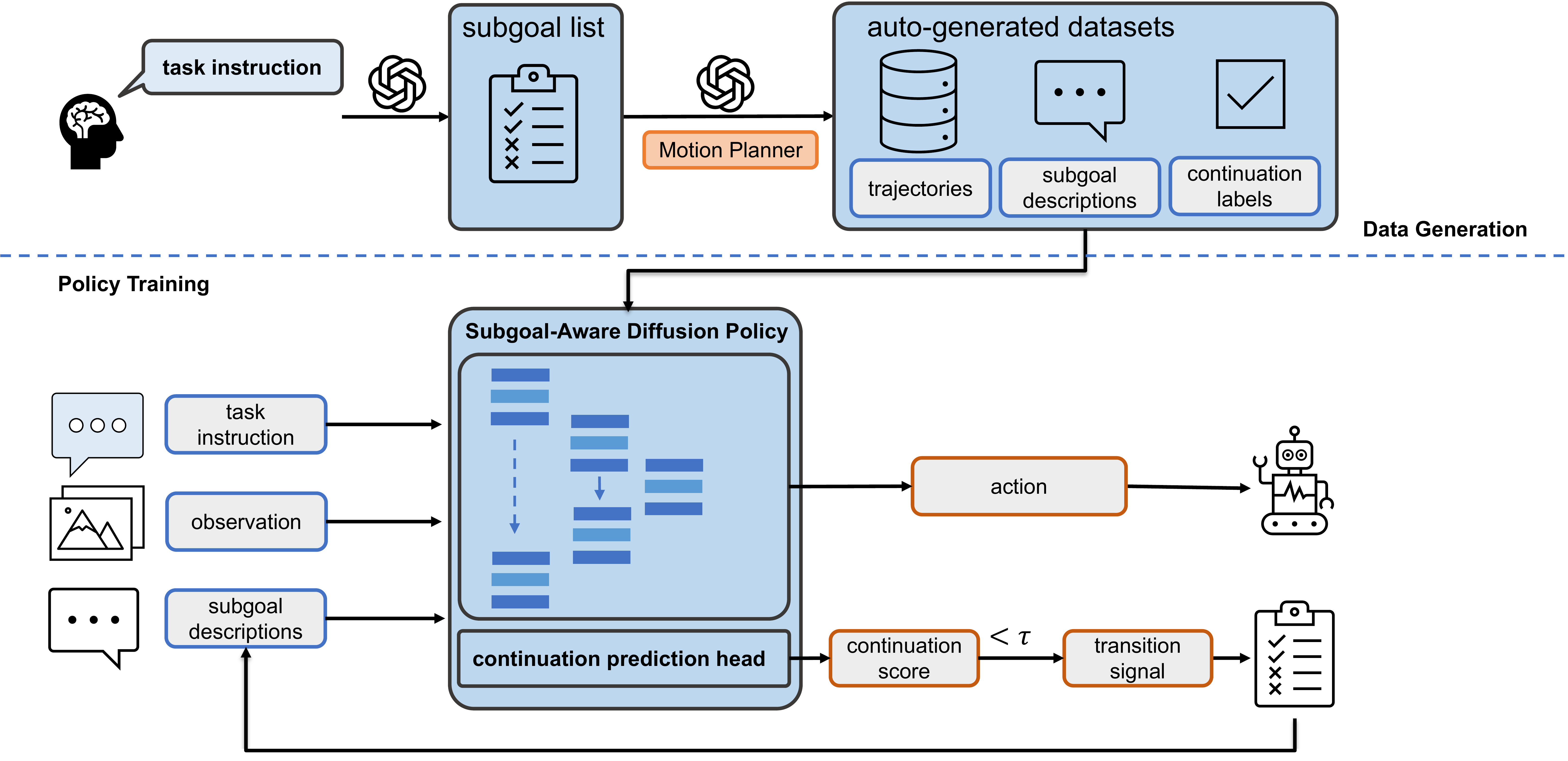}
    \caption{Overview of SADP. SADP autonomously collects demonstrations paired with structured subgoal descriptions and continuation labels to train a subgoal-aware diffusion policy with a continuation prediction head.}
    \label{fig:concept}
\end{figure}

To bridge this gap, we propose \textbf{SADP} (\textbf{S}ubgoal-\textbf{A}ware \textbf{D}iffusion \textbf{P}olicy), a framework for long-horizon robot manipulation. As illustrated in Fig.~\ref{fig:concept}, following the demonstration-generation pipeline of TARAD~\cite{hu2025tarad}, SADP leverages foundation models to decompose a task instruction into an ordered subgoal list, and executes these subgoals sequentially using an LLM-generated motion planner. Crucially, the generated subgoals are retained as natural-language execution units aligned with the automatically generated trajectory segments. Through this process, SADP autonomously generates demonstrations paired with structured subgoal descriptions and continuation labels, eliminating the need for manual subgoal annotation. Unlike prior task-level distillation approaches, SADP retains the generated subgoal descriptions and their trajectory boundaries as explicit supervision for both action generation and online subgoal transitions. Using this enriched dataset, SADP learns a diffusion-based policy that conditions action generation on both task-level instructions and subgoal descriptions, and includes a continuation prediction head that predicts a continuation score for the active subgoal. Consequently, at inference time, the active subgoal and continuation score make the policy's current execution stage explicit.

Our contributions can be summarized as follows:
\begin{itemize}
    \item We propose an automatic approach that retains the natural-language subgoals and trajectory boundaries produced during foundation-model-based demonstration generation as subgoal-level policy supervision, without manual annotation.
    \item We develop a subgoal-aware diffusion policy that conditions action generation on the active subgoal and predicts a continuation score for online subgoal switching.
    \item We evaluate SADP on simulated and real-world long-horizon manipulation tasks, demonstrating competitive task performance and explicit online estimates of the active subgoal and continuation score, and validate the continuation predictor against generator-defined subgoal boundaries on held-out demonstrations.
\end{itemize}

\section{Related Works}

\subsection{Long-Horizon Robot Manipulation}

Long-horizon manipulation tasks can typically be decomposed into a temporally extended sequence of subgoals~\cite{kroemer2021review}. HRL coordinates high-level subgoals or options with low-level policies to improve exploration, credit assignment, and temporal abstraction~\cite{hutsebaut2022hierarchical,nachum2018data,li2022hierarchical,li2023hierarchical}. These methods employ representations ranging from learned latent goals to task-specific state abstractions.

Recent IL methods explicitly model subgoal progression during execution. Jain et al.~\cite{jain2025enabling} learn explicit or attention-based transitions over visual subgoals extracted from demonstrations. SeqVLA~\cite{yang2025seqvla} augments $\pi_0$~\cite{BlackK-RSS-25} with a lightweight completion-detection head that shares features with the action expert, enabling joint action generation and autonomous transitions between subtasks. It is trained on separately collected subtask demonstrations with manually annotated binary execution/completion labels. SCIL~\cite{ouyang2025scil} combines a recurrent stage observer trained on stage-labeled demonstrations with an ACT-based stage-conditioned policy~\cite{Zhao-RSS-23}. In contrast, SADP retains the natural-language subgoals and trajectory boundaries already produced during foundation-model-based demonstration generation, using them to supervise both subgoal-conditioned action generation and continuation prediction without additional subgoal collection or manual annotation.

\subsection{Foundation Models for Robot Manipulation}

Foundation models, including large language models (LLMs) and vision-language models (VLMs), have been widely used in robot manipulation for high-level planning and visual grounding. LLM-based approaches translate natural-language instructions into high-level plans, motion primitives, or controller calls~\cite{ahn2022can,lin2023text2motion, huang2022language}, while VLM-based perception improves scene understanding and spatial grounding for language-conditioned manipulation~\cite{huang2023voxposer, liu2024grounding, ravi2024sam2, huang2024copa}.

Foundation models have also been used to scale data collection by generating demonstrations for policy learning~\cite{ha2023scaling, hu2025tarad, hua2024gensim2, jin2024robotgpt}. Although these methods reduce manual data collection and improve task-level performance, the resulting policies are usually trained as task-level controllers without retaining the subgoal structure used during demonstration generation. In SADP, this structure is retained as policy supervision for subgoal-conditioned action generation and subgoal switching.

\subsection{Diffusion Policies for Robot Manipulation}

Diffusion policies have recently emerged as effective generative models for robot manipulation, casting action generation as conditional denoising and naturally capturing complex, multi-modal action distributions~\cite{chi2023diffusion}. Building on this paradigm, prior work has improved diffusion policies through richer representations and conditioning mechanisms~\cite{hu2025tarad, ze20243d, ke20243d}. These methods typically condition action generation on observations and task-level instructions, leaving the active subgoal and its transition implicit. In contrast, SADP augments the diffusion-policy framework with active-subgoal conditioning and continuation prediction, enabling online switching through an ordered subgoal sequence.

\section{METHOD}

In this work, we propose SADP, a Subgoal-Aware Diffusion Policy framework for long-horizon robot manipulation. SADP uses natural-language subgoals to condition action generation and a continuation predictor to control online subgoal switching through the ordered subgoal sequence. As illustrated in Fig.~\ref{fig:concept}, the framework consists of two components: (1) a foundation-model-driven data generation pipeline that generates demonstrations aligned with subgoal descriptions and continuation labels, and (2) a subgoal-aware diffusion policy that jointly learns subgoal-conditioned action generation and subgoal continuation prediction.

\subsection{Subgoal-Annotated Demonstration Generation}
\label{sec:data_generation}

A major challenge in learning subgoal-conditioned policies lies in the lack of subgoal-level supervision. Most existing robot learning datasets provide only task-level labels and sparse success signals, which are insufficient for representing intermediate execution progress or decision states. Yang et al.~\cite{yang2025seqvla} address this issue by collecting demonstrations for each subgoal separately and manually annotating subgoal completion states for every frame. While effective, such manual annotation requires substantial human effort and does not scale well to diverse long-horizon tasks.

To obtain subgoal-level supervision automatically, we adopt the data-generation pipeline of TARAD~\cite{hu2025tarad} and retain the generated subgoal sequence and trajectory segmentation for policy training. Following the prompting structure of VoxPoser~\cite{huang2023voxposer}, we use GPT-4o (snapshot gpt-4o-2024-08-06)~\cite{hurst2024gpt}, with the temperature set to $0$, to decompose each natural-language instruction into a sequence of subgoals, each expressed as a short imperative phrase describing one manipulation step, and generate an affordance-based motion plan for each subgoal. The resulting plans are then sequentially executed by an LLM-generated heuristic motion planner. During execution, each trajectory is segmented according to the generated subgoal sequence, so that one segment contains the frames recorded while the planner executes the motion plan of a single subgoal. The final frame of each subgoal segment, corresponding to the last action that completes the subgoal, is labeled as 0, while all preceding frames are labeled as 1, indicating continuation of the current subgoal. Accordingly, the continuation-label sequence of each task-level demonstration follows the pattern $1\ldots10\,1\ldots10\,\ldots\,1\ldots10$. Finally, we provide GPT-4o with the final observation frame and the corresponding task description to assess task success. Trajectories deemed successful are then collected as a labeled dataset for training SADP.

\subsection{Subgoal-Aware Diffusion Policy with Continuation Prediction Head}

\begin{figure}[!t]
\centering
\includegraphics[width=\columnwidth, keepaspectratio]{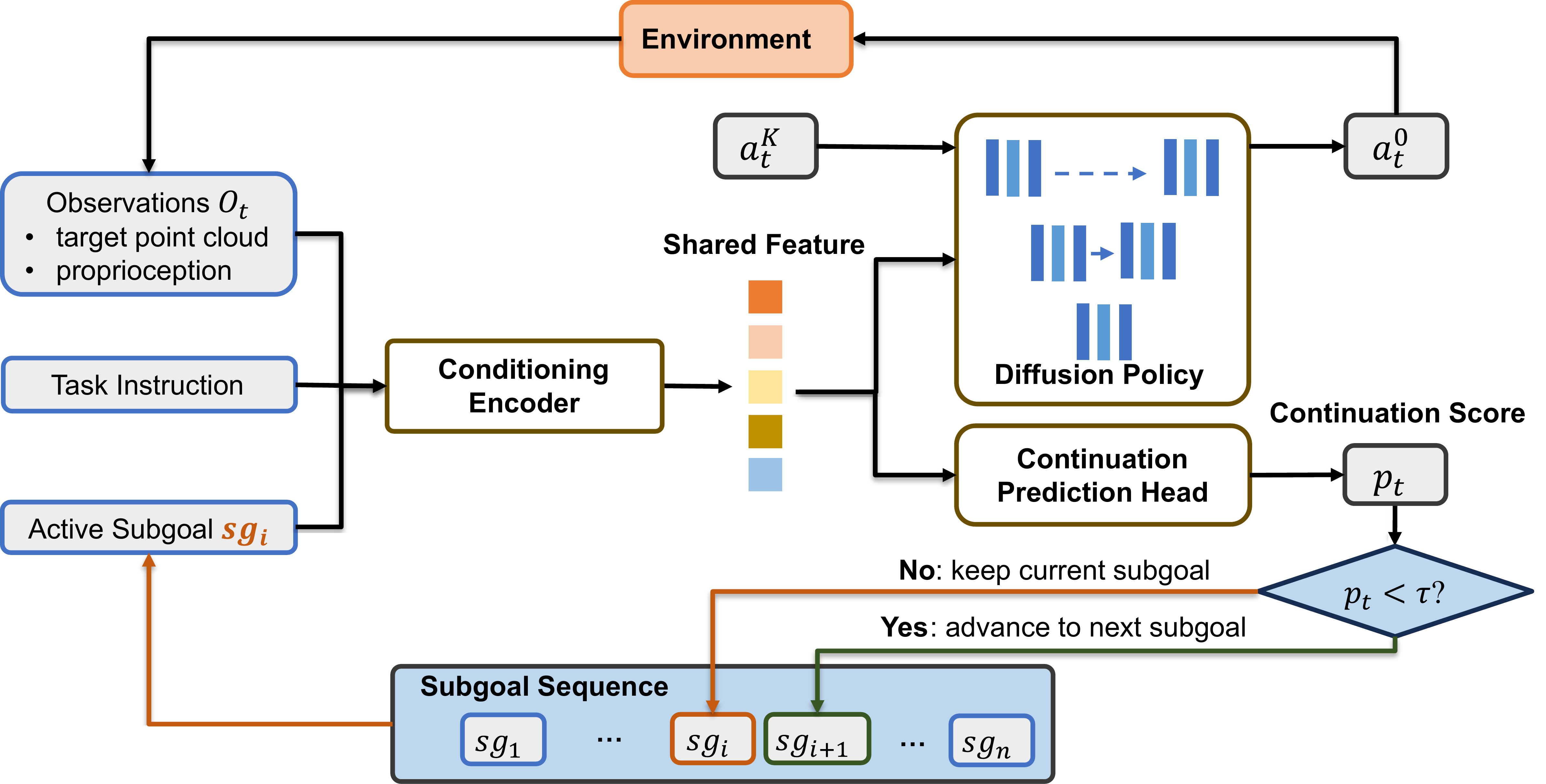}
\caption{Architecture of SADP. The current observation, task instruction, and active subgoal are encoded into a shared conditioning representation, from which the conditional diffusion policy predicts an action and the continuation head estimates the continuation score $p_t$. The resulting observation is fed back to the policy, while the active subgoal is advanced when $p_t<\tau$ and otherwise remains unchanged.}
\label{fig:framework}
\end{figure}

As shown in Fig.~\ref{fig:framework}, SADP employs a conditional diffusion policy~\cite{chi2023diffusion} that generates actions conditioned on the target-object point cloud, robot proprioceptive history, task instruction, and current subgoal description. Following the affordance-centric representation of TARAD~\cite{hu2025tarad}, the target-object point cloud is encoded using the lightweight MLP point-cloud encoder adopted from DP3~\cite{ze20243d}. The target objects of each task are identified by GPT-4o from the task description and remain fixed across subgoals. Their point clouds are extracted using the oracle segmentation masks in simulation and Grounding DINO~\cite{liu2024grounding} with SAM 2~\cite{ravi2024sam2} on the real robot, as in \cite{hu2025tarad}. The task instruction and current subgoal description are encoded using a CLIP ViT-B/32 text encoder~\cite{radford2021learning}. These representations are concatenated to form the global conditioning vector $c$, which is incorporated into the denoising network through FiLM~\cite{perez2018film}.

Starting from a random Gaussian noise vector $a_K \sim \mathcal{N}(0, I)$, the denoising network $\epsilon_\theta$ is iteratively applied for $K$ steps to obtain the final action sequence $a_0$ of horizon $H$. At each timestep $k$, the reverse diffusion update is given by

\begin{equation}
a_{k-1} = \frac{1}{\sqrt{\alpha_k}} \left( a_k - \frac{1 - \alpha_k}{\sqrt{1 - \bar{\alpha}_k}} \, \epsilon_\theta(a_k, k, c) \right) + \sigma_k z,
\end{equation} 
\noindent where $\alpha_k$ denotes the noise schedule parameter at timestep $k$, $\bar{\alpha}_k = \prod_{i=1}^{k} \alpha_i$, $\sigma_k$ is the noise scale, and $z \sim \mathcal{N}(0, I)$. At inference, we use DDIM sampling~\cite{song2021denoising}. The action prediction loss is defined as the mean squared error between the true noise $\epsilon$ and the predicted noise:

\begin{equation}
\mathcal{L}_{\text{action}} = \mathbb{E}_{a_0, \epsilon, k} \left[ \left\| \epsilon - \epsilon_\theta(a_k, k, c) \right\|^2 \right].
\end{equation}

As illustrated in Fig.~\ref{fig:framework}, SADP augments the standard conditional diffusion model with a lightweight subgoal continuation prediction head. This head is implemented as a SeqVLA-style binary classifier~\cite{yang2025seqvla} that shares the same conditioning vector $c$ with the diffusion policy and outputs a continuation score:

\begin{equation}
p_t = \sigma(W \cdot c + b),
\end{equation} 
\noindent where $\sigma(\cdot)$ denotes the sigmoid function. Since continuation frames significantly outnumber transition boundaries, we adopt focal loss~\cite{lin2017focal} to mitigate class imbalance:

\begin{equation}
\mathcal{L}_{\text{continuation}} = - \beta (1 - p)^\gamma y \log(p) - (1 - \beta) p^\gamma (1 - y) \log(1 - p),
\end{equation}

\noindent where $y \in \{0,1\}$ is the ground-truth binary label, with $y=1$ indicating continuation of the current subgoal and $y=0$ indicating a transition boundary. The focal loss hyperparameters $\beta$ and $\gamma$ are set to 0.25 and 2, respectively, following~\cite{lin2017focal}.

The overall training objective is given by
\begin{equation}
\mathcal{L}_{\text{total}} = \mathcal{L}_{\text{action}} + \lambda \, \mathcal{L}_{\text{continuation}},
\end{equation}
\noindent where $\lambda$ denotes the weight of the continuation prediction loss. It is linearly increased from $0$ to $\lambda_{max}=0.1$ over the first 1,000 training epochs and is kept fixed at $0.1$ thereafter to mitigate potential instability in early-stage action learning.

During training, we directly use the ground-truth continuation labels provided in the dataset. During inference, we generate the subgoal sequence using GPT-4o in the same manner as in the data generation stage. At each step, the policy predicts an action sequence of horizon $H$ together with the predicted continuation score $p_t$ from the current observation, and executes only the first action. After this action is executed and the environment is updated, a transition signal is issued if $p_t$ drops below a threshold $\tau$, and the policy advances to the next subgoal before the next control step. If the threshold $\tau$ is set too low, the policy may not trigger a necessary transition, whereas an excessively high threshold may cause a premature transition. We empirically set the threshold to 0.2 to balance these two effects. 

\section{EXPERIMENTS}
We evaluate SADP in both simulation and real-world environments to assess its effectiveness in long-horizon manipulation tasks. Our experiments are designed to answer the following questions:

\begin{enumerate}
    
    \item How does explicit subgoal conditioning affect task-level performance in long-horizon manipulation?
    \item How accurately and temporally consistently does the continuation predictor detect subgoal transitions?
    \item How do subgoal conditioning and transition accuracy affect long-horizon manipulation performance?
    
\end{enumerate}

\subsection{Experiment Setup}

\subsubsection{Environments}

In our simulation study, we consider a set of long-horizon manipulation tasks from RLBench~\cite{james2020rlbench}, a widely used benchmark for learning from demonstrations. These tasks require the sequential execution of multiple subgoals, posing significant challenges for long-horizon decision-making and making them particularly suitable for evaluating subgoal-aware policies.
We evaluate SADP on six representative tasks, illustrated in Fig.~\ref{fig:sim_env}. Their typical temporal structures are summarized as follows:

\begin{figure}[!ht]
    \centering
    \includegraphics[width=\columnwidth, keepaspectratio]{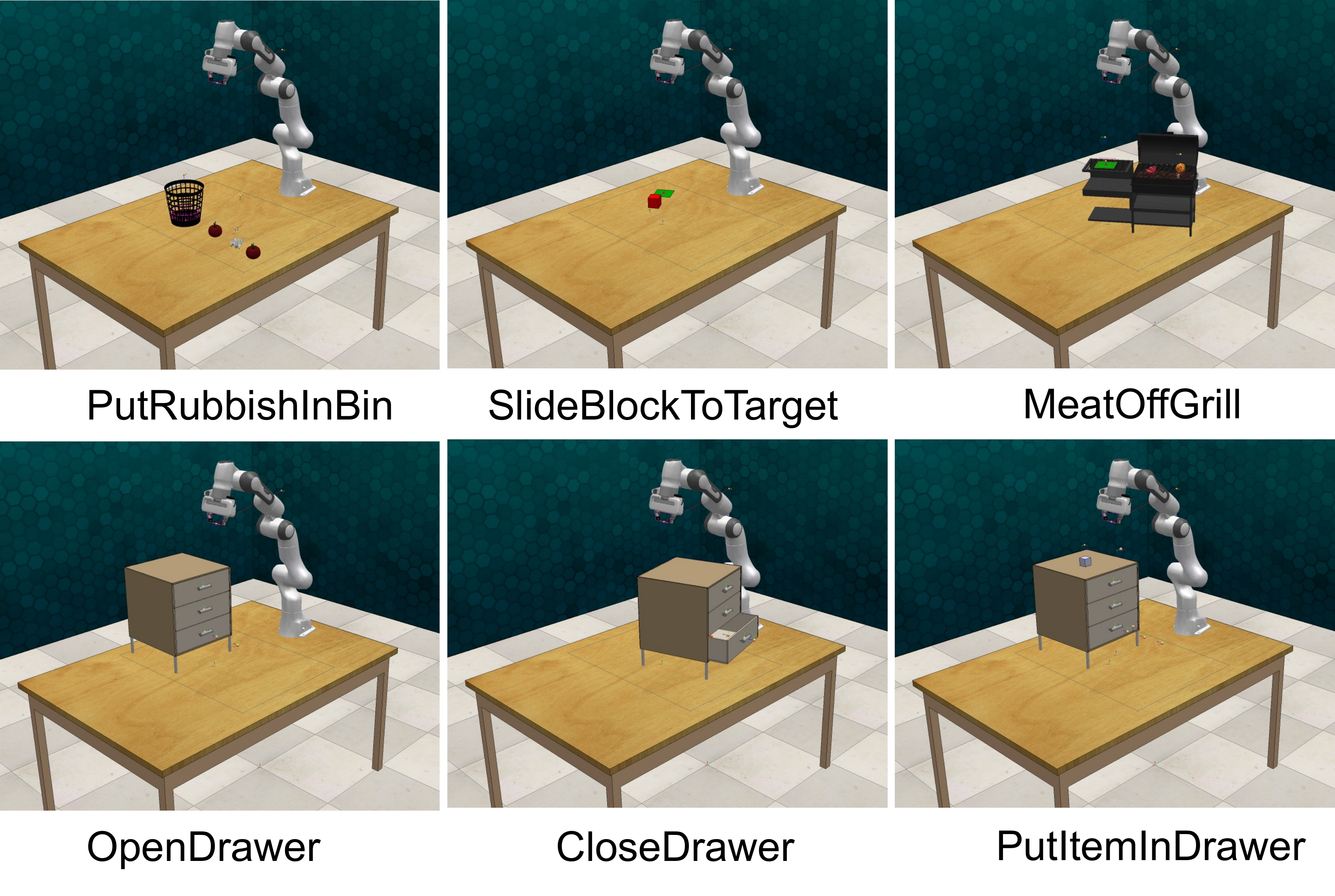}
    \caption{Six tasks in simulation experiment. The tasks cover different temporal structures, ranging from pick-and-place to drawer interaction tasks.}
    \label{fig:sim_env}
\end{figure}

\begin{enumerate}
    \item \texttt{PutRubbishInBin:} A pick-and-place task involving grasping the rubbish,  moving through an intermediate pose, reaching above the bin, and releasing it.
    \item \texttt{SlideBlockToTarget:} A pushing task involving repeated repositioning, and pushing the block toward the target zone.
    \item \texttt{MeatOffGrill:} A pick-and-place task involving grasping the target meat, moving through an intermediate pose, reaching above the target area, and releasing it.
    \item \texttt{OpenDrawer:} A drawer-interaction task involving gripper alignment, handle grasping, drawer pulling, and handle release.
    \item \texttt{CloseDrawer:} A drawer-interaction task involving gripper alignment, handle contact, and pushing the drawer closed.
    \item \texttt{PutItemInDrawer:} A longer multi-stage task combining drawer opening and object placement, involving handle manipulation, object grasping, moving above the drawer, and releasing the object inside.
\end{enumerate}

\begin{figure}[t]
    \centering
    \includegraphics[width=\columnwidth, keepaspectratio]{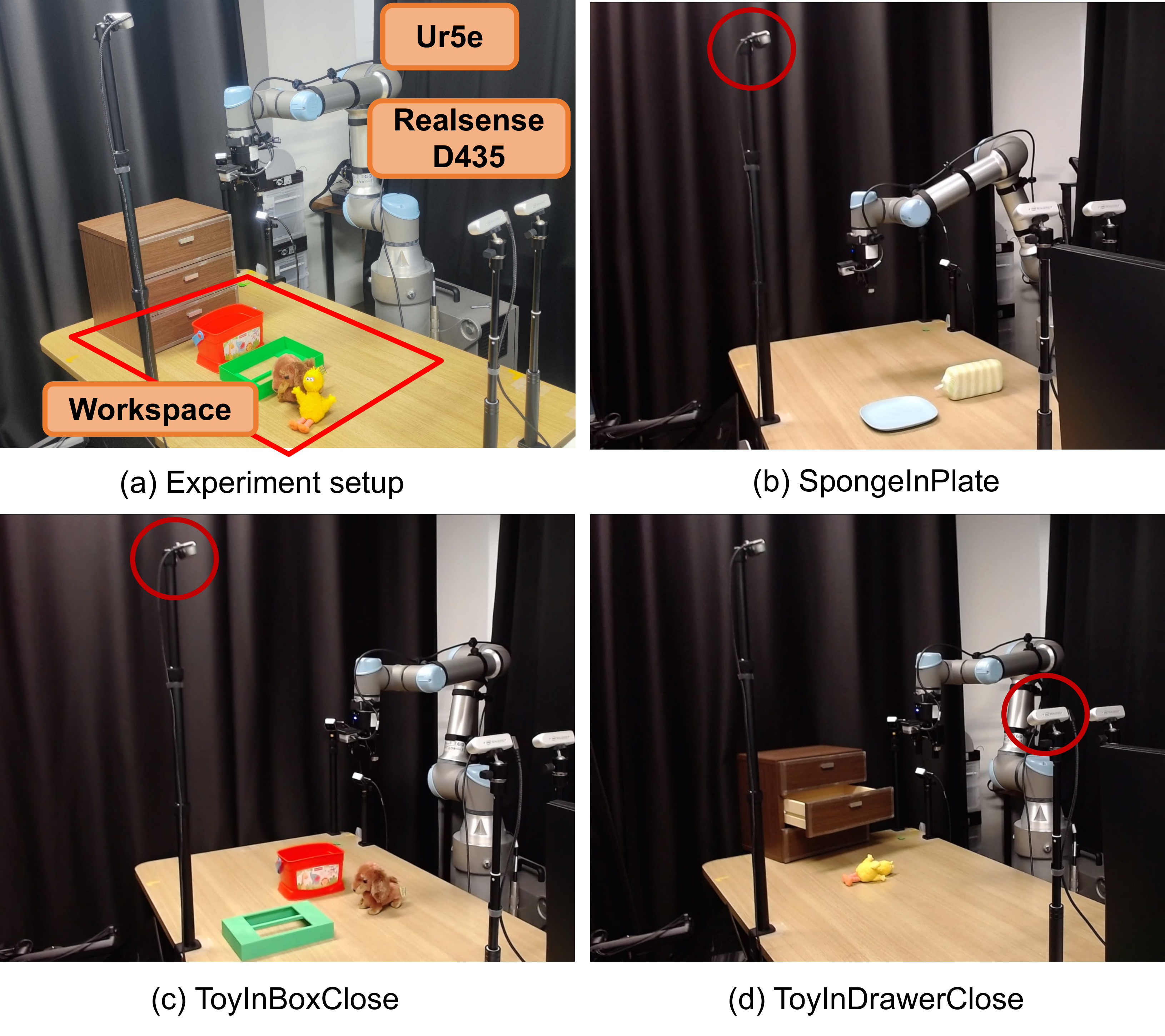}
    \caption{Real-world experiment setup. The red circles indicate the RGB-D camera used in each task. The tasks range from pick-and-place to multi-stage manipulation involving object placement followed by box or drawer closing.}
    \label{fig:real_env}
\end{figure}

We also validate SADP on real-world manipulation tasks using a UR5e robotic arm equipped with a RealSense D435 camera for RGB-D image capture. The evaluation includes three tasks, as illustrated in Fig. \ref{fig:real_env}:
\begin{enumerate}
    \item \texttt{SpongeInPlate:} A pick-and-place task involving grasping the sponge, moving through an intermediate pose, reaching above the plate, and releasing it.
    \item \texttt{ToyInBoxClose:} A longer task involving placing the toy dog into the box and then closing the lid.
    \item \texttt{ToyInDrawerClose:} A longer task involving placing the toy dog into an open drawer and then closing the drawer.
\end{enumerate}

The camera was calibrated to the UR5e base frame using the Tsai--Lenz eye-to-hand calibration method. During both demonstration generation and evaluation, the drawer was placed on the right side of the approximately $90\,\mathrm{cm}\times50\,\mathrm{cm}$ workspace, while the other objects were randomly placed within the workspace. A fixed set of task-instruction variants was generated using GPT-4o~\cite{hurst2024gpt} and shared across all methods. The same RGB-D observations, camera calibration, robot control interface, object-placement randomization, and reset protocol were used for all methods. The corresponding subgoal sequences for SADP were generated as described in Sec.~\ref{sec:data_generation}.

The control loop operated at approximately 0.9 Hz, with 250~ms for perception, 100~ms for policy inference and 750~ms for robot motion execution. Each rollout was manually terminated upon task completion, an unrecoverable failure, or reaching the 150-step limit, after which the robot and task objects were manually reset before the next rollout.

\subsubsection{Baselines}
In simulation, we compare SADP against three baselines trained on the same auto-collected datasets: a task-conditioned variant of point-cloud-based diffusion policy DP3~\cite{ze20243d} operating on scene point clouds, 3D Diffuser Actor~\cite{ke20243d} using multi-view RGB-D images from the front, left-shoulder, right-shoulder, and wrist cameras, and an affordance-centric diffusion policy TARAD~\cite{hu2025tarad} operating on target-object point clouds, a representation that SADP also adopts. In the real-world experiments, DP3 and TARAD are evaluated under the same single-camera setup as SADP. These baselines condition action generation only on observations and task-level instructions, without explicit active-subgoal conditioning or learned online switching.

\subsubsection{Dataset Construction}

Using the procedure described in Sec.~\ref{sec:data_generation}, we collected 30 automatically generated task-level demonstrations for each task, from which the observations of each method were rendered according to its input interface. The number of generation attempts per task is given in Table~\ref{tab:sim_results}. DP3, 3D Diffuser Actor, and TARAD used the task-level demonstrations, whereas SADP used the same trajectories segmented at the subgoal boundaries, together with the corresponding subgoal descriptions and continuation labels.

\subsubsection{Evaluation Metric}
For simulation experiments, we trained each task for 2000 epochs using three random seeds. For each seed, we evaluated the final checkpoint over 20 episodes per task, and report the mean and standard deviation of the success rates across the three seeds. For the real-world experiments, we trained a single seed for 2000 epochs and evaluated the final checkpoint over 30 episodes per task. Object placements were randomized independently for each trial and were not paired across methods.

\subsubsection{Implementation Details}
The diffusion policy backbone and point-cloud encoder follow DP3~\cite{ze20243d}, with all policy parameters trained from scratch. The target-object point cloud is downsampled to 512 points using farthest point sampling (FPS) and encoded into a 64-dimensional feature. The task instruction and subgoal description are separately encoded by a frozen CLIP ViT-B/32 encoder and projected from 512 to 128 dimensions using separate MLPs. The proprioceptive state is 29-dimensional. Table~\ref{tab:hyperparameters} summarizes the main hyperparameters. We set $N_{\mathrm{act}}=1$, such that the continuation score and active subgoal are updated after each executed action.

\begin{table}[t]
\centering
\caption{List of Hyperparameters of SADP.}
\label{tab:hyperparameters}
\begin{tabular}{lc}
\toprule
Setting & Value \\
\midrule
Observation horizon $N_{\mathrm{obs}}$ & 2 \\
Prediction horizon $H$ & 4 \\
Executed actions $N_{\mathrm{act}}$ & 1 \\
Diffusion scheduler & DDIM \\
Diffusion steps (training/inference) & 100 / 10 \\
Batch size & 128 \\
Optimizer / learning rate & AdamW / $1\times10^{-4}$ \\
Learning-rate schedule & Cosine \\
Training epochs & 2,000 \\
continuation-loss weight $\lambda_{max}$ & 0.1 \\
\bottomrule
\end{tabular}
\end{table}

\subsection{Results}

\begin{table*}[t]
\centering
\caption{Task success rates (\%) of SADP and baseline methods on six simulated long-horizon manipulation tasks (mean $\pm$ std over three seeds).}
\label{tab:sim_results}
\resizebox{\textwidth}{!}{
\begin{tabular}{lccccccc}
\toprule
Model & \makecell{PutRubbish\\InBin} & \makecell{SlideBlock\\ToTarget} & \makecell{MeatOff\\Grill} & \makecell{Open\\Drawer} & \makecell{Close\\Drawer} & \makecell{PutItem\\InDrawer} & \makecell{Average} \\
\midrule
Retained/Attempted (\%) & 30/40 (75.0) & 30/45 (66.7) & 30/65 (46.2) & 30/129 (23.3) & 30/30 (100.0) & 30/72 (41.7) & -- \\
\midrule
DP3~\cite{ze20243d} & 81.7 $\pm$ 2.9 & 65.0 $\pm$ 10.0 & 53.3 $\pm$ 7.6 & 55.0 $\pm$ 10.0 & 78.3 $\pm$ 5.8 & 45.0 $\pm$ 8.7 & 63.1 \\
3D Diffuser Actor~\cite{ke20243d} & 81.7 $\pm$ 10.4 & 71.7 $\pm$ 5.8 & 65.0 $\pm$ 5.0 & 61.7 $\pm$ 7.6 & 81.7 $\pm$ 2.9 & 40.0 $\pm$ 8.7 & 66.9 \\
TARAD~\cite{hu2025tarad} & 91.7 $\pm$ 5.8 & 73.3 $\pm$ 12.6 & 71.7 $\pm$ 7.6 & 71.7 $\pm$ 11.6 & 93.3 $\pm$ 2.9 & 61.7 $\pm$ 14.4 & 77.2 \\
SADP (Ours) & \textbf{93.3} $\pm$ 7.6 & \textbf{75.0} $\pm$ 10.0 & \textbf{73.3} $\pm$ 12.6 & \textbf{73.3} $\pm$ 16.1 & \textbf{95.0} $\pm$ 5.0 & \textbf{65.0} $\pm$ 18.0 & \textbf{79.2} \\
\bottomrule
\end{tabular}
}
\end{table*}

\begin{table}[t]
\centering
\caption{Main results for three real-world long-horizon manipulation tasks. Each task is evaluated with 30 trials using a single trained checkpoint; brackets give 95\% Wilson confidence intervals (\%).}
\label{tab:real_results}
\renewcommand{\arraystretch}{1.1}
\begin{tabular}{lccc}
\toprule
Model & SpongeInPlate & ToyInBoxClose & ToyInDrawerClose \\
\midrule
DP3 & \makecell{25/30\\{\scriptsize[66.4, 92.7]}} & \makecell{12/30\\{\scriptsize[24.6, 57.7]}} & \makecell{14/30\\{\scriptsize[30.2, 63.9]}} \\
TARAD & \makecell{\textbf{29/30}\\{\scriptsize[83.3, 99.4]}} & \makecell{17/30\\{\scriptsize[39.2, 72.6]}} & \makecell{23/30\\{\scriptsize[59.1, 88.2]}} \\
SADP (Ours) & \makecell{27/30\\{\scriptsize[74.4, 96.5]}} & \makecell{\textbf{22/30}\\{\scriptsize[55.6, 85.8]}} & \makecell{\textbf{26/30}\\{\scriptsize[70.3, 94.7]}} \\
\bottomrule
\end{tabular}
\end{table}

\subsubsection{Simulation Results}

Table~\ref{tab:sim_results} summarizes the simulation results. SADP achieves the highest average success rate and outperforms the task-conditioned baselines DP3 and 3D Diffuser Actor on all six tasks. The largest improvement is observed on \texttt{PutItemInDrawer}, which involves the longest subgoal sequence. Compared with TARAD, which shares the same data-generation pipeline and target-object representation but without subgoal-aware structure, SADP performs comparably overall (79.2\% vs.\ 77.2\% on average). Explicit subgoal conditioning and online subgoal switching thus preserve the task-level performance of the underlying affordance-centric policy, while making the execution stage explicit during long-horizon manipulation.

\subsubsection{Real-World Results}
Table~\ref{tab:real_results} reports the observed success counts and the corresponding 95\% Wilson confidence intervals of SADP and the baselines in real-world experiments. SADP remains effective in real-world settings. Although TARAD slightly outperforms SADP on the SpongeInPlate task, SADP outperforms both baselines on the other two tasks. Sample real-world executions are included in the supplementary video.

\subsection{Continuation Prediction}

We evaluate the continuation predictor on 30 held-out demonstrations per task, generated by the same pipeline and excluded from training, at $\tau = 0.2$. Transition boundaries ($y=0$) are treated as the positive class. We report balanced accuracy, precision, recall, and F1 score for continuation prediction, along with exact, early, and missed rates for transition timing. The boundary-detection metrics are computed over all frames in the 30 demonstrations. Timing is evaluated per subgoal segment: the annotated boundary is the final ($y=0$) frame of each segment, and a transition is triggered at the first frame whose continuation score drops below $\tau$. A segment is counted as exact if the trigger coincides with the boundary, early if it precedes the boundary, and missed if no trigger occurs within the segment. Because evaluation ends at the annotated boundary, the Missed category does not distinguish a trigger that would occur later from one that never occurs.

\begin{table}[t]
\centering
\caption{Continuation prediction results (\%) on 30 held-out demonstrations per task.}
\label{tab:continuation_results}
\scriptsize
\setlength{\tabcolsep}{2pt}
\renewcommand{\arraystretch}{0.95}
\resizebox{\columnwidth}{!}{
\begin{tabular}{lcccccccc}
\toprule
& & \multicolumn{4}{c}{Boundary Detection}
& \multicolumn{3}{c}{Transition Timing} \\
\cmidrule(lr){3-6}\cmidrule(lr){7-9}
Task & Seg. & BAcc & P & R & F1 & Exact & Early & Missed \\
\midrule
PutRubbishInBin    & 150 & 99.7 & 100.0 & 99.3 & 99.7 & 99.3 & 0.0 & 0.7 \\
SlideBlockToTarget & 257 & 81.3 & 81.9  & 72.4 & 76.9 & 66.5 & 9.3 & 24.1 \\
MeatOffGrill       & 150 & 91.3 & 93.3  & 83.3 & 88.0 & 78.7 & 4.7 & 16.7 \\
OpenDrawer         & 180 & 82.6 & 96.7  & 65.6 & 78.1 & 63.3 & 2.2 & 34.4 \\
CloseDrawer        & 423 & 92.1 & 92.7  & 87.2 & 89.9 & 81.1 & 6.1 & 12.8 \\
PutItemInDrawer    & 302 & 94.5 & 97.1  & 89.4 & 93.1 & 87.7 & 1.7 & 10.6 \\
\midrule
Average            & --  & 90.3 & 93.6  & 82.9 & 87.6 & 79.4 & 4.0 & 16.6 \\
\bottomrule
\end{tabular}}
\end{table}

As shown in Table~\ref{tab:continuation_results}, the continuation predictor exhibits a conservative switching pattern: precision is generally higher than recall, early transitions are uncommon, and missed transitions constitute the primary error source.

\begin{figure}[t]
    \centering
    \includegraphics[width=\columnwidth, keepaspectratio]{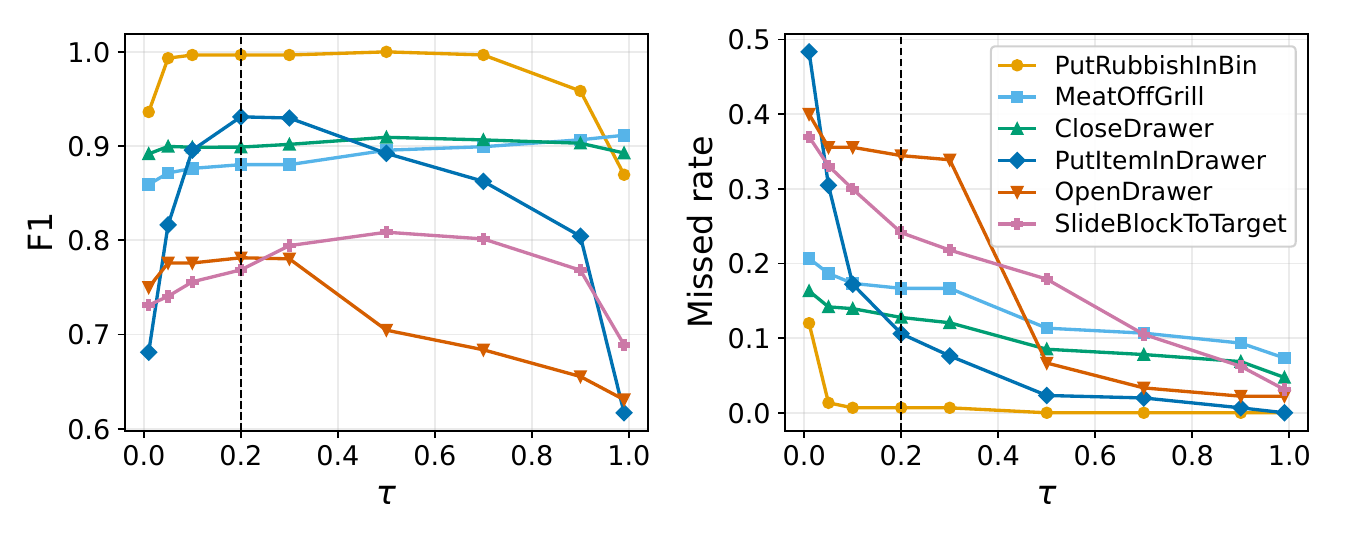}
    \caption{Threshold sensitivity across six tasks. The dashed line denotes the default threshold $\tau=0.2$.}
    \label{fig:tau_robustness}
\end{figure}

We further assess the sensitivity of continuation prediction to the transition threshold $\tau$ using the same held-out demonstrations. Figure~\ref{fig:tau_robustness} reports the F1 score and missed-transition rate as $\tau$ varies from $0.01$ to $0.99$. Performance remains relatively stable for $\tau\in[0.1,0.5]$. This robustness is consistent with the strongly bimodal distribution of continuation scores: fewer than $3.5\%$ of frames lie within $\pm 0.05$ of the default threshold. Increasing $\tau$ generally reduces the missed-transition rate, but may also decrease the F1 score by introducing additional false transitions, particularly for \texttt{PutItemInDrawer} and \texttt{OpenDrawer}.

\begin{figure}[!th]
    \centering
    \includegraphics[width=\columnwidth, keepaspectratio]{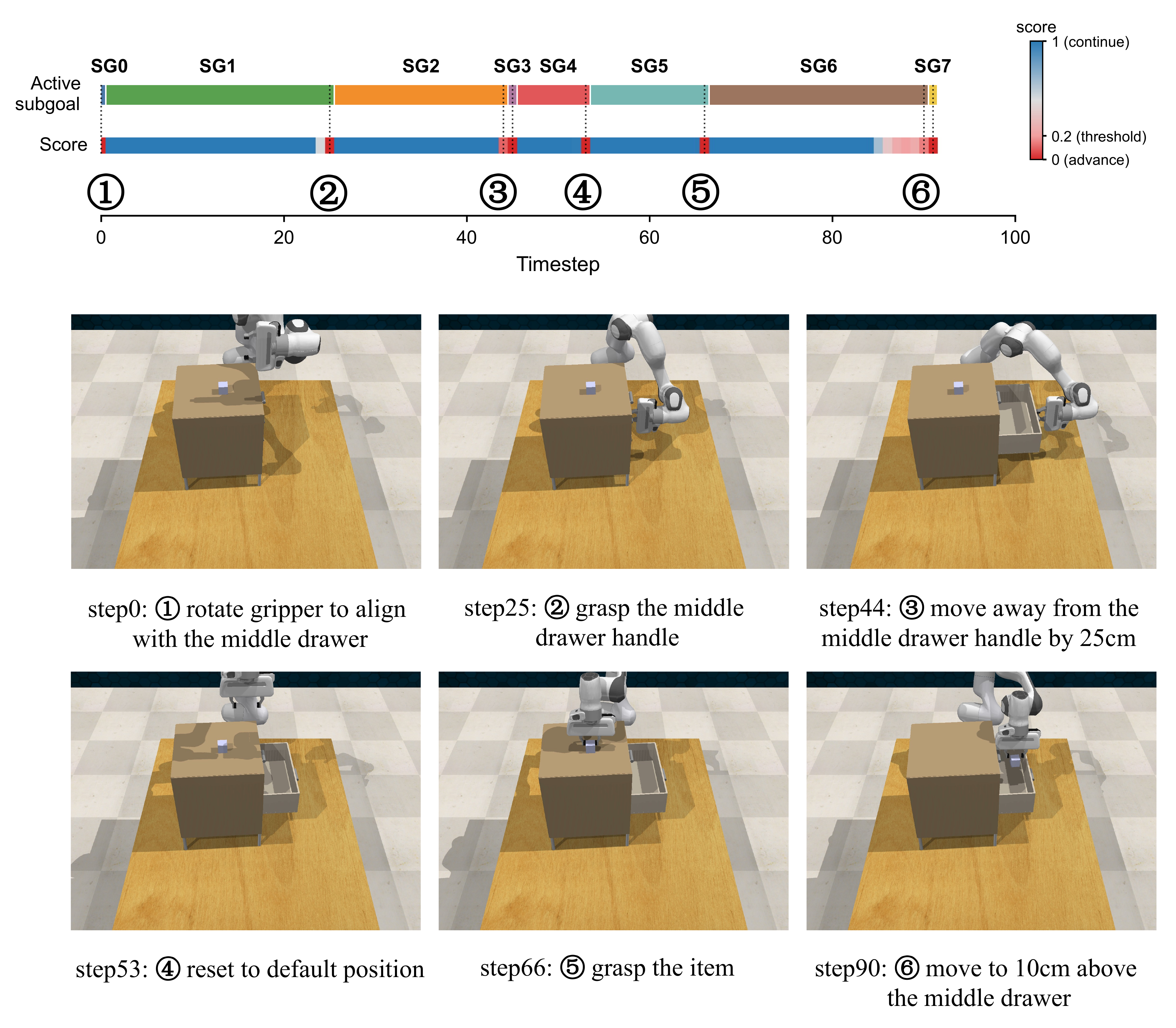}
    \caption{Predicted subgoal continuation scores for an example successful execution of \texttt{PutItemInDrawer}. The upper row shows the active subgoal over time, and the lower row shows the predicted continuation score for the current subgoal. Lower scores indicate higher confidence that the current subgoal has been completed. The predicted scores decrease around the transitions between execution stages.}
    \label{fig:success}
\end{figure}

\begin{figure}[!ht]
    \centering
    \includegraphics[width=\columnwidth, keepaspectratio]{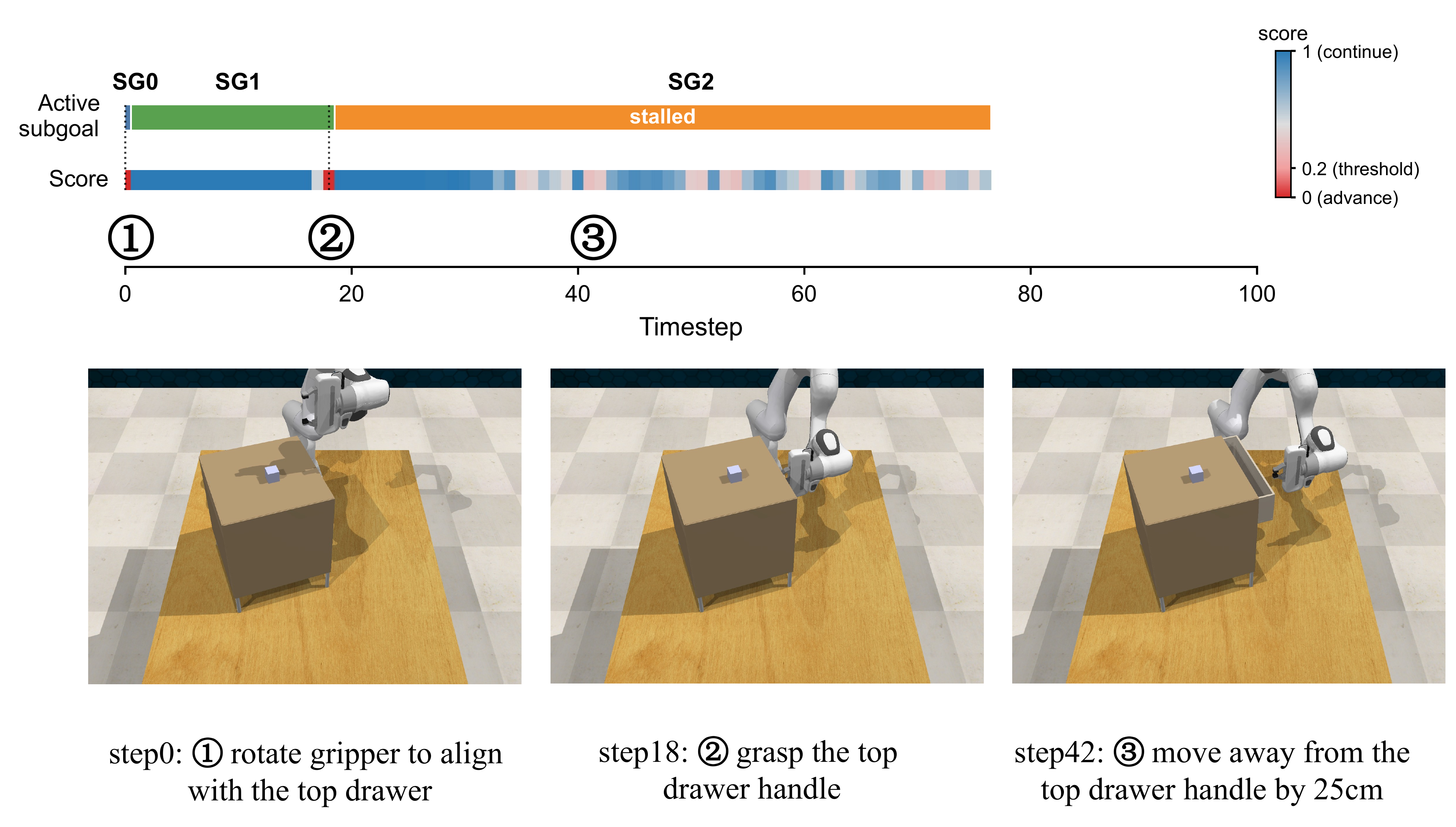}
    \caption{Predicted subgoal continuation scores for an example failed execution of \texttt{PutItemInDrawer}. The robot successfully completes the initial subgoals for aligning with and grasping the drawer handle, but the execution stalls during the drawer-opening subgoal. Because the gripper disengages from the handle during the opening motion, the drawer-opening subgoal remains incomplete, the continuation score stays above the transition threshold $\tau$, and the active subgoal does not advance. The retained active subgoal indicates the execution stage at which the robot stalls.}
    \label{fig:fail}
\end{figure}

Example online executions are shown in Figs.~\ref{fig:success} and~\ref{fig:fail}. In the successful execution, the continuation score decreases around the subgoal boundaries, and the active subgoal advances sequentially. In the failed execution, the gripper loses contact with the drawer handle during drawer opening. The continuation score remains above $\tau$, and the active subgoal remains at the drawer-opening stage, indicating where the execution stalls. These executions are included in the supplementary video.

\begin{table*}[!t]
\centering
\caption{Ablation results on subgoal conditioning, subgoal representation, and continuation prediction in simulation (mean ± std over three seeds).}
\label{tab:ablation_results}
\resizebox{\textwidth}{!}{%
\begin{tabular}{lccc|ccccccc}
\toprule
Model
& \makecell{Subgoal\\Conditioning}
& \makecell{Continuation\\Head}
& \makecell{Subgoal\\Switch}
& \makecell{PutRubbish\\InBin}
& \makecell{SlideBlock\\ToTarget}
& \makecell{MeatOff\\Grill}
& \makecell{Open\\Drawer}
& \makecell{Close\\Drawer}
& \makecell{PutItem\\InDrawer}
& Average \\
\midrule

SADP (w/o SG-Action)
& -- & $\surd$ & Pred.
& $91.7 \pm 2.9$ & $65.0 \pm 8.7$ & $55.0 \pm 8.7$
& $56.7 \pm 16.1$ & $78.3 \pm 10.4$ & $50.0 \pm 5.0$ & 66.1 \\

SADP (SG-Index)
& Index & $\surd$ & Pred.
& $88.3 \pm 2.9$ & $41.7 \pm 10.4$ & $61.7 \pm 5.8$
& $61.7 \pm 11.5$ & $88.3 \pm 2.9$ & $45.0 \pm 8.7$ & 64.4 \\

\textbf{SADP}
& Text & $\surd$ & Pred.
& $93.3 \pm 7.6$ & $75.0 \pm 10.0$ & $73.3 \pm 12.6$
& $73.3 \pm 16.1$ & $95.0 \pm 5.0$ & $65.0 \pm 18.0$ & 79.2 \\

\midrule

SADP (SG-Action Only)
& Text & -- & Manual
& $\mathbf{96.7} \pm 2.9$ & $76.7 \pm 7.6$ & $78.3 \pm 5.8$
& $75.0 \pm 8.7$ & $95.0 \pm 5.0$ & $60.0 \pm 5.0$ & 80.3 \\

SADP (Manual Transition)
& Text & $\surd$ & Manual
& $93.3 \pm 2.9$ & $\mathbf{86.7} \pm 7.6$ & $\mathbf{83.3} \pm 10.4$
& $\mathbf{80.0} \pm 5.0$ & $\mathbf{96.7} \pm 2.9$
& $\mathbf{66.7} \pm 7.6$ & \textbf{84.4} \\

\bottomrule
\end{tabular}%
}
\end{table*}

\subsection{Ablation Study} 

We evaluate four ablative versions of SADP in simulation: (i) SADP (w/o SG-Action) removes the subgoal input from the action branch, while the continuation head still receives the subgoal embedding; (ii) SADP (SG-Index) replaces each language subgoal with its ordinal index, embedded by a learned 128-dimensional embedding table; (iii) SADP (SG-Action Only) removes the continuation head and uses manual transition signals, in which a human operator determines whether the current subgoal has been completed and manually advances the policy to the next subgoal; and (iv) SADP (Manual Transition) replaces only the predicted transitions with manual signals. The results are summarized in Table~\ref{tab:ablation_results}.

Removing subgoal input from the action branch drops the task success rate below that of TARAD. We attribute this degradation to a structural mismatch: because the model is still trained to predict subgoal continuation, the shared representation is encouraged to encode subgoal progress that the action branch cannot exploit. Replacing the language descriptions with ordinal indices retains part of the benefit on tasks whose stages are behaviorally distinct, such as \texttt{OpenDrawer} and \texttt{CloseDrawer}, but fails on \texttt{SlideBlockToTarget}, whose repeated pushing stages perform the same behavior yet are assigned different indices. Language descriptions can avoid both failure modes, distinguishing behaviorally different stages while letting identical subgoals share supervision.

Manual transitions raise the average success rate from 79.2\% to 84.4\%. The task-specific improvements broadly correspond to the transition-timing accuracy reported in Table~\ref{tab:continuation_results}: the gains are largest on \texttt{SlideBlockToTarget}, \texttt{MeatOffGrill}, and \texttt {OpenDrawer}, where transition timing is frequently inaccurate, but minimal on \texttt{PutRubbishInBin}, where transitions are predicted almost perfectly. On \texttt{PutItemInDrawer}, even manual switching yields only a small gain (+1.7) and the success rate remains the lowest among all tasks, suggesting that its failures are dominated by the manipulation difficulty of its long stage sequence rather than by switching errors. Under the same manual switching, the jointly trained model outperforms SADP (SG-Action Only), suggesting that continuation supervision also contributes to improved action generation.

Finally, we assess the sensitivity to the continuation-loss weight $\lambda_{max}$ on \texttt{MeatOffGrill}, keeping the warm-up schedule and all other settings fixed. As shown in Table~\ref{tab:lambda_sensitivity}, success rates remain stable for $\lambda_{max}$ between 0.05 and 0.5, with the highest mean at 0.2 and the default 0.1 performing comparably. In contrast, $\lambda_{max} = 1.0$ leads to a decrease, suggesting that excessive weighting of the auxiliary continuation objective can impair action learning.

\begin{table}[t]
\centering
\caption{Sensitivity to the continuation-loss weight $\lambda_{max}$ on
MeatOffGrill (task success rate, \%) (mean ± std over three seeds).}
\label{tab:lambda_sensitivity}
\resizebox{\columnwidth}{!}{
\begin{tabular}{lccccc}
\toprule
$\lambda_{max}$ & 0.05 & 0.10$^{\dagger}$ & 0.20 & 0.50 & 1.00 \\
\midrule
Success Rate & 71.7 $\pm$ 7.6 & 73.3 $\pm$ 12.6 & \textbf{75.0} $\pm$ 8.7
& 70.0 $\pm$ 13.2 & 63.3 $\pm$ 2.9 \\
\bottomrule
\multicolumn{6}{l}{\footnotesize $^{\dagger}$Default.}
\end{tabular}}
\end{table}

\section{DISCUSSION}

By conditioning the policy on the current natural-language subgoal, SADP exposes the current execution stage while maintaining performance comparable to the strongest task-conditioned baseline in simulation and achieving higher success counts on two of the three real-world tasks. The subgoal structure produced during demonstration generation remains useful beyond data generation: retaining it during policy learning helps organize long-horizon execution around explicit intermediate stages. Moreover, because only successful generated trajectories are retained, the resulting subgoal sequences are associated with executable robot behavior.

The ablation results further show that the subgoal descriptions do more than expose the execution stage: they also serve as an effective conditioning input for action generation, outperforming ordinal stage indices. The main remaining limitation is the accuracy of continuation prediction: subgoals occupy unequal durations in the collected demonstrations, so short subgoals provide few samples near their transition boundaries, and the learned predictor switches conservatively; the resulting timing errors are a main source of the gap to manual switching. Our evaluation of the continuation predictor is also limited to offline agreement with generator-defined boundaries on held-out demonstrations; its stage alignment during closed-loop execution remains to be assessed. Future work will investigate temporally informed, continuous progress estimation beyond binary continuation prediction, together with an online stage-annotation study and a systematic attribution of execution failures to transition versus control errors.

\section{CONCLUSIONS}

In this work, we proposed SADP, a subgoal-aware diffusion policy trained from demonstrations automatically generated by foundation models. By decomposing long-horizon instructions into ordered natural-language subgoals and executing them sequentially, SADP eliminates the need for manual annotation of subgoal descriptions and continuation labels. Through the integration of subgoal conditioning and a continuation prediction head, SADP embeds explicit natural-language subgoal structure directly into diffusion-based action generation.

Experimental results in simulation and real-robot settings show that SADP maintains competitive task performance while exposing the active subgoal and its continuation score during execution, and its continuation predictor recovers most generator-defined subgoal boundaries in offline evaluation. The manual-transition results further point to transition prediction as an important bottleneck in long-horizon execution. Overall, these findings demonstrate that subgoal-level execution transparency can be integrated into a diffusion policy while retaining competitive control performance.






\bibliographystyle{template/IEEEtran} 
\bibliography{template/IEEEabrv.bib, root.bib} 

\end{document}